%% file: acl.tex
\pdfoutput=1

\documentclass[11pt]{article}

\usepackage[]{acl}
\usepackage{multirow}
\usepackage{graphicx}
\usepackage{booktabs}
\usepackage{tabularx}
\usepackage{CJKutf8}
\usepackage{amssymb}
\usepackage{amsmath}
\usepackage{tabu} 
\usepackage{hyperref}
\usepackage{breakurl}
\usepackage{times}
\usepackage{latexsym}
\usepackage{tikz}
\usepackage{pgfplots}
\usepackage[normalem]{ulem}
\usepackage{xcolor}
\usepackage{dashrule}

\definecolor{brickred}{HTML}{b92622}
\definecolor{midnightblue}{HTML}{005c7f}
\definecolor{salmon}{HTML}{f1958d}
\definecolor{burntorange}{HTML}{f19249}
\definecolor{junglegreen}{HTML}{4dae9d}
\definecolor{forestgreen}{HTML}{499c5e}
\definecolor{pinegreen}{HTML}{3d8a75}
\definecolor{seagreen}{HTML}{6bc1a2}
\definecolor{limegreen}{HTML}{97c65a}

\usepackage{lipsum}

\newcommand\blfootnote[1]{%
  \begingroup
  \renewcommand\thefootnote{}\footnote{#1}%
  \addtocounter{footnote}{-1}%
  \endgroup
}

\usepackage[T1]{fontenc}

\usepackage[utf8]{inputenc}

\usepackage{microtype}

\title{
MuCGEC: a Multi-Reference Multi-Source 
 Evaluation Dataset for \\ Chinese Grammatical Error Correction}

\author{Yue Zhang$^{1\dagger}$, Zhenghua Li$^{1*}$, Zuyi Bao$^2$, Jiacheng Li$^1$, \\ {\bf Bo Zhang$^2$},   {\bf Chen Li$^2$}, {\bf Fei Huang$^2$}, {\bf Min Zhang$^1$} \\
        $^1$Institute of Artificial Intelligence, School of Computer Science and Technology, \\
Soochow University, China; $^2$DAMO Academy, Alibaba Group, China\\
\texttt{$^1$\{yzhang21,jcli20\}@stu.suda.edu.cn}, \texttt{$^1$\{zhli13,minzhang\}@suda.edu.cn}\\\texttt{$^2$\{zuyi.bzy,klayzhang.zb,puji.lc,f.huang\}@alibaba-inc.com}}

\begin{document}
\begin{CJK}{UTF8}{gkai}

\maketitle

\input{chapters/abstract}
\input{chapters/introduction_new}

\input{chapters/data_annotation}

\input{chapters/data_analysis}

\input{chapters/baselines}

\input{chapters/experiments}

\input{chapters/related_work}

\input{chapters/conclusion}

\bibliography{anthology,custom}
\bibliographystyle{acl_natbib}

\clearpage

\input{chapters/appendix}

\end{CJK}
\end{document}

%% file: chapters/abstract.tex
\begin{abstract}

This paper presents MuCGEC, a multi-reference multi-source evaluation dataset for Chinese Grammatical Error Correction (CGEC), consisting of 7,063 sentences collected from three Chinese-as-a-Second-Language (CSL) learner sources. Each sentence is corrected by three annotators, and their corrections are carefully reviewed by a senior annotator, resulting in 2.3 references per sentence. We  conduct experiments with two mainstream CGEC models, i.e., the sequence-to-sequence model and the sequence-to-edit model, both enhanced with large pretrained language models, achieving competitive benchmark  performance on  previous and our datasets. We also discuss CGEC evaluation methodologies, including the effect of multiple references and using a char-based metric. Our annotation guidelines, data, and code are available at \url{https://github.com/HillZhang1999/MuCGEC}.

\end{abstract}

%% file: chapters/introduction_new.tex
\section{Introduction}
\label{sec:intro}

Given a potentially noisy input sentence, grammatical error correction (GEC) aims to detect and correct all errors and produce a clean sentence.
Recently, GEC has increasingly gained attention for its vital value in various downstream scenarios \citep{grundkiewicz2020crash, wang2021comprehensive}.\blfootnote{$^\dagger$ This work was partially done during the first author's internship at Alibaba DAMO Academy.}\blfootnote{$^*$ Corresponding author.}

To support GEC research, high-quality manually labeled evaluation data is indispensable. 
For English GEC (EGEC), such datasets are abundant \citep{yannakoudakis2011new, dahlmeier2013building, ng2014conll,napoles2017jfleg,bryant2019bea,napoles2019enabling,flachs2020grammatical}. 
However, Chinese GEC (CGEC) evaluation datasets are relatively scarce. 
The two
publicly available CGEC evaluation datasets are NLPCC18 and CGED, contributed by the NLPCC-2018 \citep{zhao2018overview} and the series of CGED shared tasks \citep{rao2018overview,rao2020overview}, respectively.

\input{tables/gec_example}

Most %
EGEC evaluation datasets provide multiple references for each input sentence, such as CoNLL14-test  \citep{ng2014conll} and BEA19-test \citep{bryant2019bea}. In contrast, sentences in existing CGEC evaluation datasets usually have only one reference (i.e., 87\% of the sentences in NLPCC18 and all in CGED). This is probably due to the different annotation workflows adopted. 

As suggested by \citet{bryant2015far},
enforcing multi-reference annotation is crucial for both GEC model evaluation and GEC data annotation, 
because there usually exist more than one acceptable reference with similar meanings for an incorrect sentence, as illustrated by the example in Table \ref{tab:example}. 
On the one hand, if a GEC model outputs a correct reference, which is yet different from the one given in the evaluation data, 
then the model performance will be unfairly underestimated. 
To mitigate this issue, a straightforward solution is  increasing the number of references  \citep{sakaguchi2016reassessing, choshen2018inherent}.
\textcolor{black}{
On the other hand, imposing a single-reference constraint makes data annotation problematic. 
If annotators submit different equally acceptable corrections, which is very common, it will be difficult for the senior annotator to decide which one is the best.
}

Besides the lack of multiple references, existing CGEC  datasets collect sentences from a single text source,  which may be insufficient for robust model evaluation  \citep{mita2019cross}. 
In addition, we believe that it is beneficial for  improving data quality to compile comprehensive annotation guidelines.

To fill these gaps, this paper presents a multi-reference multi-source evaluation dataset for CGEC, named MuCGEC.  
After investigating previous works on constructing GEC datasets, we compile comprehensive annotation guidelines.
Based on a browser-based online annotation tool, each  sentence is assigned to three annotators for independent correction, and one senior annotator for final review. 
An annotator may submit multiple references, and 
the senior annotator may also supplement new references besides rejecting incorrect submissions.
In this way, we aim to produce as many references as possible. 
In summary, this work makes the following contributions. %
\begin{itemize}
    \item [(1)] Our newly constructed MuCGEC 
    consists of 7,063 sentences from three representative sources of Chinese-as-a-Second-Language (CSL) learner texts. 
    Each sentence obtains 2.3 references on average. 
    We conduct detailed analyses on our new dataset to  gain more insights. %
    \item [(2)] We conduct benchmark experiments using two mainstream and competitive CGEC models, i.e., the sequence-to-edit (Seq2Edit) and sequence-to-sequence (Seq2Seq) models, both enhanced with pretrained language models (PLMs).
    We also experiment with an extremely effective  ensemble strategy.
    Moreover, we investigate the effect of multiple references on model evaluation, and propose to use a char-based evaluation metric, which we believe is  simpler and more suitable than previous word-based ones for CGEC. 
\end{itemize}

%% file: tables/gec_example.tex
\begin{table}[]
\scalebox{0.93}{
\begin{tabular}{ll}
\toprule
\textbf{Source} & \begin{tabular}[c]{@{}l@{}}我不知道他何时返回回来。\\ I don't know when he will return back.\end{tabular} \\ \hline
\textbf{Ref. 1} & \begin{tabular}[c]{@{}l@{}}我不知道他何时返回\sout{回来}。\\ I don't know when he will return.\end{tabular}        \\ 
\textbf{Ref. 2} & \begin{tabular}[c]{@{}l@{}}我不知道他何时\sout{返回}回来。\\ I don't know when he will be back.\end{tabular}       \\ 
\bottomrule 
\end{tabular}
}

\caption{A CGEC example %
with two references.%
}
\label{tab:example}
\end{table}

%% file: chapters/data_annotation.tex
\section{Data Annotation}

\subsection{Multi-Source Data Selection}

This work focuses on CSL learner texts. In order to investigate diverse types of Chinese grammatical errors, we select data from the following three sources. 
\begin{itemize}
\item [(1)] We re-annotate the NLPCC18 test set \citep{zhao2018overview}, which contains 2,000 sentences from the Peking University (PKU) Chinese Learner Corpus.
\item [(2)] We select and re-annotate sentences from CGED-2018 and CGED-2020 test datasets \citep{rao2018overview,rao2020overview}, which come from the writing section of the HSK exam (Hanyu Shuiping Kaoshi, translated as the Chinese level exam), an official Chinese proficiency test. After removing sentences marked as correct from the total 5,006 ones,  we obtain 
3,137 potentially erroneous sentences for re-annotation. %

\item [(3)] Lang8\footnote{\url{https://lang-8.com/}} is a language learning platform, where native speakers  voluntarily correct texts written by second-language  learners. 
The NLPCC-2018 shared task organizers collect about 717K Chinese sentences with their corrections
from Lang8 and encourage participants to use them as the training data.
We randomly select 2,000 sentences with 30 to 60  characters for re-annotation.
\end{itemize} 

In the end, we obtain 7,137 sentences.
For simplicity, we discard all original corrections and directly perform re-annotation from scratch following our new annotation guidelines and workflow.

\subsection{Annotation Paradigm: Direct Rewriting} 

There are mainly two types of annotation paradigms for constructing GEC data, i.e., %
\emph{error-coded} and \emph{direct rewriting}. 
The \emph{error-coded} paradigm requires annotators to explicitly mark the erroneous span in the original sentence, then choose its error type, and finally make corrections. 
\citet{ng2013conll,ng2014conll} adopt the \emph{error-coded} paradigm for constructing data for the CoNLL-2013/2014 EGEC shared tasks. 
For CGEC, the original NLPCC18 and CGED datasets both follow the \emph{error-coded} paradigm as well.

As discussed by \citet{sakaguchi2016reassessing}, the \emph{error-coded} paradigm poses two  challenges. First, it is extremely difficult for different annotators to agree upon the boundaries of the erroneous spans and their error types, especially when there are many categories to consider \citep{bryant2017automatic}. This inevitably leads to an increase in annotation effort and a decrease in annotation quality. Second, under such a complex annotation paradigm, annotators would pay less attention to the fluency of the resulting reference, sometimes even leading to unnatural expressions.

Instead, the \emph{direct rewriting} paradigm asks annotators to directly rewrite the input sentence and produce a corresponding grammatically correct one, without changing the original meaning. 
In order to evaluate model performance, edits can be extracted automatically from parallel sentences by additional tools \citep{bryant2017automatic}. 

This annotation paradigm has proven to be efficient and cost-effective \citep{sakaguchi2016reassessing}, and has been adopted by several recent GEC data construction works 
\citep{napoles2017jfleg,napoles2019enabling,syvokon2021ua,10.1162/tacl_a_00470}.

In this work, we adopt the \emph{direct rewriting} paradigm. Besides above-mentioned advantages, we believe this paradigm can help improve the diversity of references since annotators can correct errors more freely.  

\subsection{Annotation Guidelines}
\input{tables/error_categories_guideline}

After an extensive survey of previous work on GEC data construction, we compiled 30 pages of comprehensive guidelines for CGEC annotation. During the course of the annotation process, we gradually improved our guidelines according to feedback from annotators.

To facilitate learning, our guidelines adopt a two-tier hierarchical error taxonomy,
including 5 major error types and 14 minor types, as shown in Table \ref{tab:error:category:guideline}. 
Our guidelines describe in detail how to handle each minor error type and provide typical examples. 
We will release our guidelines along with the dataset, %
which we hope can benefit future research.

For dealing with word-missing errors,  we found that it was unreasonable to simply insert certain words when the missing words are context-dependent, which means the missing words are related to context beyond the given sentence. 
Table~\ref{tab:mc} shows a sentence in which a verb is missing.  
However, under sentence-level GEC annotation, annotators are unable to decide the specific missing verb. 
According to our observation, previous CGEC datasets directly insert
specific words like ``scolds'' under such circumstances, which we think is inaccurate and may cause trouble for GEC model evaluation, because there are many other acceptable candidates.
To handle this problem, we instead insert a special tag named  \textbf{context-dependent missing components (MC)}.
We find about 1\% of sentences in MuCGEC contain ``[MC]'' tags. 
Current GEC models cannot handle ``[MC]'', since ``[MC]'' is not included in existing training data and vocabulary. We leave this issue as future work.  

\input{tables/mc}

\input{tables/overall_statistic}
\subsection{Annotation Workflow and Tool}

In order to encourage more diverse and high-quality references, we assign each sentence to three random annotators for independent annotation.
\textcolor{black}{Their submissions are then aggregated and sent to a random senior annotator (reviewer) for review.}
An annotator may submit multiple references for one sentence if he/she  thinks they are all correct according to the guidelines. 
The job of the senior annotator includes: 1) modifying incorrect  references into correct ones (sometimes just rejecting them); 2) adding new correct references according to the guidelines. 
After review, the accepted references are defined as \textbf{final golden references}. %

For the sake of self-improvement, we employ a self-study mechanism that allows annotators to learn from their mistakes if they submit an incorrect reference. 
Concretely, if an annotator submits a reference that is not included in the final golden references, he/she has to modify his/her submission into a correct one. 
Moreover, the annotator can also make complaints if he/she insists that his/her submission is correct. 
We find that the self-study and making-complaints mechanisms can trigger very helpful discussions.

To improve annotation efficiency, we have developed 
 a browser-based online annotation tool to support the above workflow and mechanisms. Due to the space limitation, we show the visual interfaces for annotation and review in Appendix \ref{sec:interface}.

\subsection{Annotation Process}
\label{section:qc}

We employed 21 undergraduate students who are native speakers of Chinese and familiar with Chinese grammar as part-time annotators. Annotators received intensive training %
before real annotation. 
In the beginning, two authors of this paper, who were also in charge of  compiling the guidelines, 
served as senior annotators for review. After one month, when the  annotators were familiar with the job, we selected 5 outstanding annotators as senior annotators to join the review. 

All participants were asked to annotate for at least 1 hour every day. The whole annotation process lasted for about 3 months.%

\subsection{Ethical Issues}

All annotators and reviewers were paid for their work. The salary was  determined by both submission numbers and annotation quality. The average salary of annotators and reviewers is 24 and 35 RMB per hour  respectively. 

All the data of the three sources are publicly available. Meanwhile, %
we have obtained permission from organizers of the NLPCC-2018 and CGED shared tasks to release our newly annotated references in a proper way. 
We are deeply grateful to them for their kind support.

%% file: tables/error_categories_guideline.tex
\begin{center}
\begin{table}[tb]
\centering
\scalebox{0.85}{
\begin{tabular}{@{}lp{6cm}@{}}
\toprule
\textbf{Major Types} & \textbf{Minor Types}                                                   \\ \midrule
\multirow{1}{*}{Punctuation}                   & Missing; Redundancy; Misuse \\ \midrule
\multirow{1}{*}{Spelling}                      & Phonetic confusion; Glyph confusion; Character disorder \\ \midrule
\multirow{1}{*}{Word}                          & Missing; Redundancy; Misuse                      \\ \midrule
\multirow{1}{*}{Syntax}                        & Word order;  Mixing syntax patterns \\ \midrule
\multirow{1}{*}{Pragmatics}                    & Logical inconsistency; Ambiguity; Commonsense mistake \\ \bottomrule
\end{tabular}
}
\caption{The 5 major and 14 minor error types adopted by our guidelines for  organizing the content.
}
\label{tab:error:category:guideline}
\end{table}
\vspace{-0.5cm}
\end{center}

%% file: tables/mc.tex
\begin{table}[]
\centering
\begin{tabular}{cc}
\hline
\multirow{2}{*}{\textbf{Source}}      & 我的爸爸经常我。                    \\
                                      & My dad usually me.           \\ \hline
\multirow{2}{*}{\textbf{Previous}} & 我的爸爸经常\uline{骂}我。                   \\
                                      & My dad usually \uline{scolds} me.   \\ 
\multirow{2}{*}{\textbf{Ours}}   & 我的爸爸经常\uline{{[}MC{]}}我             \\
                                      & My dad usually \uline{{[}MC{]}} me. \\ \hline
\end{tabular}
\caption{An example for handling context-dependent missing components. The inserted tokens are underlined. ``Previous'' means annotation in previous datasets, and ``Ours'' refers to our annotation.}
\label{tab:mc}
\end{table}

%% file: tables/overall_statistic.tex
\begin{center}
\begin{table*}[!htbp]
\centering
\begin{tabular}{llccccc}
\toprule
\textbf{Dataset} & \textbf{\#sent} & \textbf{\#err. sent (perc.)} & \textbf{chars/sent} & \textbf{edits/ref} & \textbf{refs/sent} \\ \hline
Original NLPCC18 & 2000              & 1983 (99.2\%)             & 29.7               & 2.0                & 1.1                             \\ \hline
MuCGEC (NLPCC18)     & 1996 ~(~4)             & 1904 (95.4\%)            & 29.7               & 2.5                & 2.5                           \\
MuCGEC (CGED)        & 3125 (12)             & 2988 (95.6\%)           & 44.8               & 4.0                & 2.3                              \\
MuCGEC (Lang8)       & 1942 (58)              & 1652 (85.1\%)           & 37.5               & 2.8                & 2.1                             \\ 
\hline
MuCGEC         & 7063 (74)              & 6544 (92.7\%)           & 38.5               & 3.2                & 2.3                             \\ \bottomrule
\end{tabular}
\caption{Data statistics, including the number of sentences, the number (proportion) of erroneous sentences, the average number of characters per sentence, the average number of edits per reference, and the average number of references per sentence.
Some sentences in our source data were discarded since annotators could not understand their meaning and thus were unable to correct them. 
Numbers in parentheses in the ``\#sent'' row refer to such sentences. 
}
\label{tab:overall:statistic}
\end{table*}
\vspace{-0.5cm}
\end{center}

%% file: chapters/data_analysis.tex
\section{Analysis of Our Annotated Data}

\textbf{Overall statistics} of MuCGEC are shown in Table  \ref{tab:overall:statistic}. 
We also include the original NLPCC18 dataset \citep{zhao2018overview} for comparison\footnote{
Here we do not compare with the original CGED and Lang8 datasets since: 1) the CGED-orig mainly focuses on error detection annotation and does not provide corrections for word-order errors; 2) the Lang8-orig is automatically collected from the internet, and its correction is quite noisy.}.

First, from the proportion of erroneous sentences, 
we can see that most of the sentences are considered to contain  grammatical errors in the original annotation, but a considerable part of them are not corrected in our annotation. We attribute this to our strict control of the over-correction phenomenon.

\label{sec:ana}
Second, regarding sentence lengths, NLPCC18 has the shortest sentences, whereas CGED sentences are much longer.
This may be because candidates on the HSK examination, an official Chinese proficiency test, tend to use long sentences to show their ability in Chinese.

Third, each sentence in the re-annotated NLPCC18 receives 2.5 references on average, which is more than twice that in the original NLPCC18 data. 
Overall, each sentence obtains 2.3 references. 
We believe the multi-reference characteristic makes our dataset more reliable for evaluation, which is further discussed in Section \ref{sec:6.3}.

\label{sec:edit}

Finally, we compare the number of char-based edits per reference in different datasets. We describe how to derive such edits in detail in Section \ref{sec:exp}.
We can see that the number of edit is tightly correlated with sentence length. The difference in the average sentence length and number of edits indicates that the three data sources may have a systematic discrepancy in quality and difficulty, which we believe is helpful for evaluating the generalization ability of models.
Moreover, compared with NLPCC18-orig, we annotate 25\% more edits (2.0 vs. 2.5) in each reference. We believe the major reason is that the original NLPCC18 data are annotated under the \emph{minimal edit distance}  principle \citep{nagata2016phrase}, which requires annotators to select a reference
\textcolor{black}{with fewer edits when correcting errors.} %

\input{figure/reference_proportion}

\textbf{Sentence distribution with respect to numbers of references} 
is shown in Figure \ref{fig:ref:dist}. 
Here, we only consider erroneous sentences. Identical references from different annotators are counted as one reference.
Overall, most sentences have 2 references, accounting for 39.4\%; 29.1\% of sentences have 3 references; 21.8\% of sentences have only 1 reference, most of which are short and easy to correct.

We believe that the average number of references could be further increased if more annotators were assigned to each sentence.  
It is also worth noticing that annotators tend to submit a single reference, despite the fact that our annotation tool allows annotators to submit multiple ones. %
We suspect the reason may be that it is more economical for annotators to do so. One the one hand, 
it may be easy to come up with the most suitable correction, whereas thinking of alternatives is more time-consuming.
On the other hand, we did not give enough consideration to this issue when designing the salary computation formula. 
In the future, we plan to optimize (or simplify) our annotation workflow  so that each annotator is required to give only one reference which he/she thinks is the best, and assign each sentence to more annotators if we need more references.

\textbf{Human annotation performance.}
In order to assess the annotation ability of our annotators and human performance for CGEC task, 
\textcolor{black}{we calculate char-based F$_{0.5}$ scores by evaluating all annotation submissions against the final golden references.
}
We describe how to compute char-based metrics in detail in Section \ref{sec:exp}. 
Each reference submitted by an annotator is considered as a sample. 
Overall, the average F$_{0.5}$ is 72.12, which we believe could be higher if we discarded data that were annotated at the early stage of our project when annotators were less experienced and less familiar with our guidelines.

Figure \ref{fig:annotator:f} shows F$_{0.5}$ scores of 15 annotators who annotated the most sentences, in descending order of the number of annotated sentences.
We can see that human performance varies across different annotators. The best annotator achieves an 82.34 F$_{0.5}$ score,
while the annotator who completes the most tasks only gets a score of 68.32.
\textcolor{black}{This indicates that we should pay more attention to annotation quality when calculating salaries and prevent annotators from focusing too much on annotation speed.}
\input{figure/annotator_f0.5}

\textbf{Common mistakes made by annotators.
}
\textcolor{black}{We randomly select 300 invalid references rejected by reviewers and try to understand what mistakes are more frequently made by annotators. %
} 
We manually classify all selected references into three mistake  categories.
The most frequent \textcolor{black}{mistakes} are caused by \emph{incomplete correction} and account for 56.7\% of the invalid references. We found that it was sometimes difficult for annotators to correct all errors without omissions, possibly due to the complexity or flexibility of Chinese grammar. \textcolor{black}{The second frequent type of mistakes is }\emph{erroneous correction}, which means that the correction of old errors incurs new errors, with a proportion of 32.3\%. Besides, 11.0\% of submissions are rejected due to \emph{meaning change}, which means that the correction changes the intended meaning of the original sentence.

%% file: figure/reference_proportion.tex
\usetikzlibrary{patterns}

\begin{figure}[tb!] %
\centering %
\scalebox{0.7}{

\begin{tikzpicture}
\begin{axis}[
ybar,
xlabel=Number of references,
ylabel=Proportion of sentences (\%),
ymin=0,
ytick = {0, 10,..., 40},
symbolic x coords={1,2,3,4,$\ge5$},,
xtick=data,
nodes near coords,
nodes near coords align={vertical},
ymajorgrids=true,
grid style=dashed,
width=9.5cm,
]
\addplot [draw=black, pattern=horizontal lines light blue
] coordinates {(1,21.80) (2,39.4) (3,29.1) (4,7.6) ($\ge5$, 2.1) };
\end{axis}
\end{tikzpicture}

}

\caption{The proportion of sentences with different number of references in MuCGEC.} 

\label{fig:ref:dist}
\end{figure}
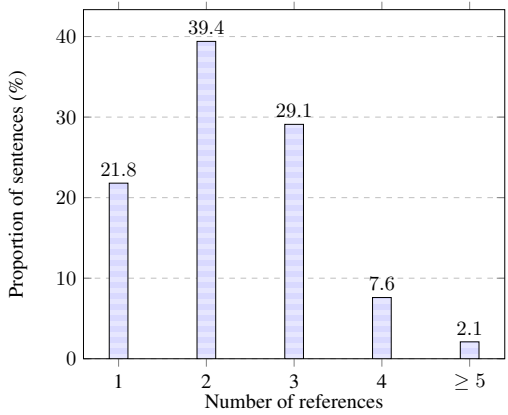

%% file: figure/annotator_f0.5.tex
\begin{figure}[tb!] %
\centering %
\scalebox{0.7}{

\begin{tikzpicture}
\begin{axis}[
	ylabel=$F_{0.5}$,
	xlabel=Annotator ID,
	ymin=61.5,
	ymax=82.5,
	enlargelimits=0.02,
	ybar interval=0.6,
	ymajorgrids=true,
	xmajorgrids=false,
    grid style=dashed,
    width=9.5cm,
]
\addplot [draw=black, pattern=horizontal lines light blue
]
	coordinates {(1,68.32) (2,71.25) (3,73.04) (4,68.75) (5,75.07) (6,76.13) (7,67.01) (8,69.42)(9,76.85) (10,70.67) (11,71.91) (12,62.06) (13,66.26) (14,82.34) (15,77.99) (16,79.30)};
\end{axis}
\end{tikzpicture}

}

\caption{The human performance of the 15 annotators who annotated the most sentences.}

\label{fig:annotator:f}
\end{figure}
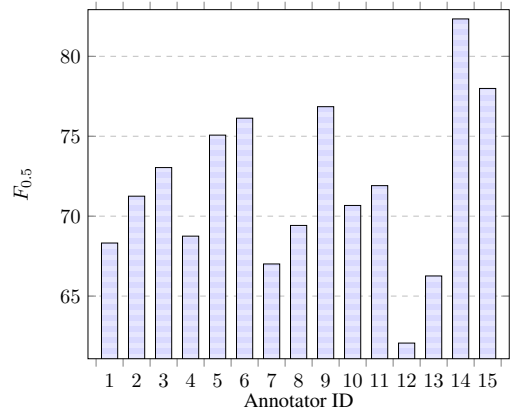

%% file: chapters/baselines.tex
\section{Benchmark Models}
\label{sec:bench}

To understand how well cutting-edge GEC models perform on our data, we adopt two mainstream GEC approaches, i.e., Seq2Edit and Seq2Seq.  %
Both models are enhanced with PLMs. We also attempt to combine them after observing their complementary power in dealing with different error types. 
This section briefly describes these benchmark models. Due to the space limitation, please kindly refer to Appendix \ref{sec:hp} for more model details.

\textbf{The {Seq2Edit} model} 
treats GEC as a sequence labeling task and performs error corrections via a sequence of token-level edits, including insertion, deletion, and substitution \citep{malmi2019encode}. A token corresponds to a word or a subword in English, and to a character in  Chinese. With minor modifications to accommodate Chinese, we adopt GECToR \citep{omelianchuk2020gector}, which achieves the SOTA performance on EGEC datasets. 

Following recent Seq2Edit work like  \citet{awasthi2019parallel} and \citet{omelianchuk2020gector}, 
we enhance GECToR by using PLMs as its encoder. 
After comparing several popular PLMs, we 
choose StructBERT \citep{wang2019structbert}\footnote{\url{https://github.com/alibaba/AliceMind/tree/main/StructBERT}}
due to its superior performance after fine-tuning
(see Table \ref{tab:nlpcc18:res}).

\textbf{The Seq2Seq model}  
straightforwardly treats GEC as a monolingual translation task \citep{yuan2016grammatical}. 
Recent works propose to enhance Transformer-based \citep{vaswani2017attention} Seq2Seq EGEC models with PLMs like T5 \citep{rothe2021recipe} or BART \citep{katsumata2020stronger}. 
Unlike BERT \citep{devlin2019bert}, T5 and BART are specifically designed for text generation. Therefore, it is straightforward to continue training them on GEC data. 
\textcolor{black}{%
We follow these works and utilize the recently proposed Chinese BART from \citet{shao2021cpt} to initialize our Seq2Seq model.}

\textbf{The ensemble model.} Several previous works have proven the effectiveness of model ensemble for CGEC \citep{liang2020bert,hinson2020heterogeneous}. In this work, we clearly observe the complementary power of the above two models in fixing different error types (see Table \ref{tab:type:eval}), and thus attempt to combine them. 

We adopt a simple edit-wise voting mechanism. The edits are at the char-based span level, and correspond to four error types. Please refer to Section \ref{sec:eval-metric} for detailed explanation of our char-based evaluation metric.  
More specifically, we aggregate edits from the results of each model, and only preserve edits that appear more than $N/2$ times, where $N$ is the number of models. In other words, an edit is kept in the final result only if it is produced by a majority of models.\footnote{ 
In fact, our voting strategy is a little more complex due to the adaption of two pieces of modification, which consistently improve performance in our preliminary experiments. 
First, for word-order errors, we set the preserving threshold to $N/2-1$
considering the Seq2Seq model is much superior in handling word-order errors than the Seq2Edit model. Imagine the scenario when all Seq2Seq models agree on correcting a word-order error, whereas all Seq2Edit models disagree. 
Using $N/2-1$ means that a word-order edit is kept in the final result even when it is produced by exactly half of models.
Second, we use a set of simple heuristic rules to recognize spelling  errors, a sub-type of substitution errors, and also use $N/2-1$ as the preserving threshold for them. 
This is also reasonable since both GEC models can obtain high precision scores on such errors. 
}

We experiment with two ensemble settings: 1) one Seq2Edit and one Seq2Seq, denoted by ``1$\times$Seq2Edit+1$\times$Seq2Seq'', and 2) three Seq2Edit and three Seq2Seq, denoted by ``3$\times$Seq2Edit+3$\times$Seq2Seq''.The three Seq2Edit models are obtained using different random seeds for initialization, and the same goes for the Seq2Seq.

\textbf{Other settings.} To obtain the single-model performance of both kinds of models, we run them three times separately with different random seeds for initialization and calculate average metrics.
For ``1$\times$Seq2Edit+1$\times$Seq2Seq'', we random select a pair of single models. 
For ``3$\times$Seq2Edit+3$\times$Seq2Seq'', we aggregate the results of all six single models.

\section{Experiments on NLPCC18-Orig Data}

In order to show that our benchmark models are competitive among existing CGEC models, we conduct experiments on the original NLPCC18 test set, on which most previous CGEC systems are tested. 

\textbf{Training data.} 
For the sake of easy replicability, 
we limit our training data strictly to public resources, i.e., the Lang8\footnote{\burl{http://tcci.ccf.org.cn/conference/2018/taskdata.php}} data \citep{zhao2018overview} and the HSK\footnote{\burl{http://hsk.blcu.edu.cn}} data \citep{zhang2009hsk}.
We filter duplicate sentences that appear in our dataset, 
and discard correct sentences.
The final Lang8 and HSK data contains 1,092,285 and 95,320 sentence pairs, respectively. The HSK data is cleaner and of higher quality than Lang8, but is smaller. Following the re-weighting procedure of  \citet{junczys2018approaching}, we duplicate the HSK data five times, and merge them with Lang8 data.

\input{tables/nlpcc2018_results}
\textbf{Comparison with previous works.}
Table \ref{tab:nlpcc18:res} shows the results.
For a fair comparison, we follow the official setting of the shared task, including the word-based MaxMatch scorer  \citep{dahlmeier2012better} for calculating the P/R/F values.
We segment model outputs by adopting the PKUNLP word segmentation (WS) tool provided by the shared task organizers \citep{zhao2018overview}.

\label{sec:4}

When only using Lang8 for training, our single Seq2Seq model is already quite competitive. Its performance is only lower than MaskGEC \citep{zhao2020maskgec} by 1 point in F$_{0.5}$.
Please notice that MaskGEC extra uses data augmentation. 
For now, our benchmark models do not use any synthetic data for simplicity, but 
we believe data augmentation could further boost the performance of our models. 

Adding the HSK training data improves performance of all our models  %
by about 4 points. %
Our two benchmark models already achieve SOTA performance under the  single-model setting.

The model ensemble technique leads to obvious performance gains (more than 5 points) over single models. 
However, the gains from increasing the number of component models seem rather small. 
We try to explain this issue in Section \ref{sec:mucgec-results-ana}. 

For Seq2Edit, we additionally present results  with other PLMs besides StructBERT, including 
BERT \citep{devlin2019bert},
RoBERTa \citep{liu2019roberta},
and MacBERT \citep{cui2020revisiting} from the Hugging Face\footnote{\url{https://huggingface.co/}} website.
We use the ``large'' variants of all PLMs.

%% file: tables/nlpcc2018_results.tex
\begin{table}[]
\scalebox{0.75}{
\begin{tabular}{lccc}
\toprule
& \textbf{P}               & \textbf{R}               & \textbf{$F_{0.5}$}           \\ \hline
\multicolumn{4}{c}{\textbf{Trained on Lang8}} \\
YouDao \citep{fu2018youdao}$\diamondsuit$ &35.24          & 18.64          & 29.91          \\
AliGM \citep{zhou2018chinese}$\diamondsuit$                   & 41.00          & 13.75          & 29.36          \\
BLCU \citep{ren2018sequence}$\diamondsuit$                     & 47.63          & 12.56          & 30.57          \\
HRG \citep{hinson2020heterogeneous}$\diamondsuit$                     & 36.79          & \textbf{27.82}          & 34.56          \\
MaskGEC \citep{zhao2020maskgec}$\heartsuit$                  & \textbf{44.36} & 22.18          & \textbf{36.97}          \\
\hline
\emph{Our Seq2Edit} & \textbf{39.83}          & 23.01          & 34.75          \\
\emph{Our Seq2Seq} & 37.67          & \textbf{29.88}          & \textbf{35.80}          \\
\hline
\emph{1$\times$Seq2Edit+1$\times$Seq2Seq$\diamondsuit$} & \textbf{58.15}          & 18.35          & 40.55          \\
\emph{3$\times$Seq2Edit+3$\times$Seq2Seq$\diamondsuit$} & 55.58          & \textbf{19.78}          & \textbf{40.81}          \\

\hline 
\hline

\multicolumn{4}{c}{\textbf{Trained on Lang8+HSK}} \\
TEA \citep{wang2020base}$\heartsuit$                  & 39.43 & 22.80          & 34.41          \\
WCDA \citep{tang2021WCDA}$\heartsuit$                  & \textbf{47.41} & \textbf{23.72}          & \textbf{39.51}          \\
\hline
\emph{Our Seq2Edit (BERT)}   & 39.61         & 28.53          & 36.76          \\
\emph{Our Seq2Edit (RoBERTa)} & 39.74          & 30.44          & 37.54          \\
\emph{Our Seq2Edit (MacBERT)} & 40.46          & {30.73}          & 38.05          \\ 
\emph{Our Seq2Edit (StructBERT)}  & \textbf{42.88 }         &30.19          &\textbf{39.55}         \\
\emph{Our Seq2Seq} & 41.44          & \textbf{32.89}          & 39.39      \\
\hline
\emph{1$\times$Seq2Edit+1$\times$Seq2Seq$\diamondsuit$} & \textbf{60.72}          & 22.48          & 45.31          \\
\emph{3$\times$Seq2Edit+3$\times$Seq2Seq$\diamondsuit$} & 59.38          & \textbf{24.18}          & \textbf{45.99}          \\
\bottomrule
\end{tabular}
}
\caption{Performance comparison on the original NLPCC18 dataset \citep{zhao2018overview} using the official \textbf{word-based} evaluation script. The first group lists models that use only Lang8 for training, whereas the second group shows those using both Lang8 and HSK data. Models marked by $\diamondsuit$ use \emph{model ensemble}, and those marked by $\heartsuit$ use \emph{data augmentation}. 
}
\label{tab:nlpcc18:res}
\end{table}

%% file: chapters/experiments.tex
\section{Experiments on MuCGEC}

\input{tables/benchmark}

\subsection{Data Splits} 
For hyperparameter tuning or model selection, previous works on other CGEC datasets often randomly  sample some sentence pairs from training data as the dev set \citep{wang2020base, zhao2020maskgec, hinson2020heterogeneous}, which is inconvenient for reproducing or comparing.

In this work, we propose to provide a fixed dev set for our newly annotated dataset, by randomly selecting 1,125 sentences from the CGED source, denoted as CGED-dev.
The remaining 5,938 sentences are used as the test set, in which each data source has a roughly equal amount of sentences, i.e., 1,996 sentences for NLPCC18-test, 2,000 for CGED-test, and 1,942 for Lang8-test.

\subsection{Evaluation Metrics} \label{sec:eval-metric}

\label{sec:exp}

\textbf{Problems with word-based metrics.} As discussed in Section \ref{sec:4}, previous CGEC datasets are annotated upon word sequences and adopt word-based metrics for evaluation. 
Before annotation and evaluation, a sentence needs to be segmented into words using a Chinese word segmentation (CWS) model. 
We believe this will introduce unnecessary uncertainty in CGEC evaluation procedure. %
First, CWS models inevitably produce word segmentation errors \citep{fu2020chinese}. Second, there are multiple heterogeneous CWS standards.  
Finally, we found that a correct edit may be judged as wrong due to the word boundary mismatch.

\input{tables/error_type_eval}

\textbf{Char-based span-level evaluation metrics} are adopted in this work instead.
First, given an input sentence and a correction, we obtain an optimal  sequence of \emph{char-level edits} with the minimal edit distance. 
We consider three types of \emph{char-level edits}, corresponding to three error types:  
\vspace{-3mm}
\begin{itemize} 
    \setlength{\itemsep}{-2pt}
    \setlength{\topsep}{-2pt}
    \item Deleting a char for a \emph{redundant error};
    \item Inserting a char for a \emph{missing error};
    \item Substituting a char with another one for a \emph{substitution error};
\end{itemize}
\vspace{-3mm}

Then, we convert all \textit{char-level edits} into \textit{span-level} by merging consecutive edits of the same type,  
following previous practice in EGEC and CGEC  \citep{felice2016automatic,hinson2020heterogeneous}

The above two steps are applied to both the system output sequence and golden reference, transforming them into sets of span-level edits.  
Finally, we can calculate the P/R/F value by comparing them. 
If there are multiple golden references, we will choose the one with the highest F-score. 

\textbf{Span-level word-order errors.} When calculating overall metrics, we only consider above three types of errors. When analyzing, we distinguish the fourth error type --- word-order. A span-level word-order error is usually composed of a redundant and a missing error, where the deleted span is the same as the inserted one. We use simple heuristic rules to identify such errors \cite{hinson2020heterogeneous}. 

Please kindly notice that we release our evaluation script as well.

\subsection{Results and Analysis}\label{sec:mucgec-results-ana}

\label{sec:6.3}
\textbf{Main results.} 
Table \ref{tab:benchmark:ours} shows  the char-based performance of the  benchmark models and our annotators 
on MuCGEC. 
All models are trained on Lang8+HSK, as described in Section \ref{sec:bench}.

The overall trend of performance change is basically consistent with those on the original NLPCC18 dataset in  Table \ref{tab:nlpcc18:res}. 
First, the Seq2Seq and Seq2Edit models perform quite closely on F$_{0.5}$, but clearly exhibit divergent strength in precision and recall, 
giving a strong motivation for combining them. 
Second, the model ensemble approach improves performance by a very large margin.  

One interesting observation is that on MuCGEC, ``3$\times$Seq2Edit+3$\times$Seq2Seq'' substantially outperforms ``1$\times$Seq2Edit+1$\times$Seq2Seq'' on All-test and all three subsets. %
In contrast, the improvement is only modest on the original NLPCC18 test data. We suspect this may indicate that a multi-reference dataset can more accurately evaluate model performance. However, it may require further human investigation for more insights.

Finally, there is still a huge performance gap between models and humans, indicating that CGEC research still has a long way to go.

\input{figure/answer_num}
\textbf{Performance on four  error types.} Table \ref{tab:type:eval} shows more fine-grained evaluation results on four error types. 

It is clear that the Seq2Edit model is better at handling redundant errors, whereas the Seq2Seq model is superior in dealing with substitution and word-order errors. 
For missing errors, the two perform similarly well. 

These phenomena are quite interesting and can be understood after considering the underlying model architectures. 
On the one hand, to correct redundant errors, the  Seq2Edit model only needs to perform a fixed deletion operation, which is a much more implicit choice for the {Seq2Seq} model, since its goal is to rewrite the whole sentence. 
On the other hand, the Seq2Seq is suitable to substitute or reorder  words due to its natural capability of utilizing language model information, 
especially with the enhancement of 
BART \citep{lewis2020bart}.

Again, the model ensemble approach substantially improves performance on all error types. The ensemble model is closest to the human on redundant errors, probably because they are the easiest to correct. The largest gap occurs in word-order errors, which require global structure knowledge to correct and are extremely challenging.

\textbf{Influence of the number of references.}
To understand the impact of the number of references on performance  evaluation, we deliberately reduce the available number of reference in our dataset. For example, when the maximum number of references is limited to 2, 
we only kept the first two references in the dataset if a sentence has more than 2 golden references. 
The results are shown in
Figure \ref{fig:answer:num}. %

When the maximum number of references increases, the performance of both models and humans increases continuously, especially for humans. 
As only a few sentences have more than 3 references, the improvement is quite small when the maximum number of references increases from 3 to All. %
This trend suggests that compared with single-reference datasets, a multi-reference dataset reduces the risk of underestimating performance, and thus is more reliable for model evaluation.

%% file: tables/benchmark.tex
\begin{table*}[!htbp]
\centering
\scalebox{0.8}{
\begin{tabular}{l|ccc|ccc|ccc|ccc}
\toprule
  & \multicolumn{3}{c|}{\textbf{NLPCC18-test}} & \multicolumn{3}{c|}{\textbf{CGED-test}} & \multicolumn{3}{c|}{\textbf{Lang8-test}} & \multicolumn{3}{c}{\textbf{All-test}}  \\ 
& \textbf{P}  & \textbf{R}  & \textbf{$F_{0.5}$}  & \textbf{P} & \textbf{R} & \textbf{$F_{0.5}$} & \textbf{P}  & \textbf{R} & \textbf{$F_{0.5}$} & \textbf{P} & \textbf{R} & \textbf{$F_{0.5}$} \\ \midrule
\textit{Seq2Edit}           & \textbf{50.09}       & 32.09       & \textbf{45.04}         & 42.87       & \textbf{27.69}      & 38.64          & \textbf{39.65}      & 21.62      & \textbf{33.98}         & \textbf{44.11}      & 27.18      & 39.22         \\
\textit{Seq2Seq}           & 47.99       & \textbf{35.12}       & 44.71         & \textbf{46.04}       & 26.97      & \textbf{40.34}          & 36.10      & \textbf{25.01}      & 33.16         & 43.81      & \textbf{28.56}      & \textbf{39.58}         \\ \midrule
\textit{1$\times$Seq2Edit+1$\times$Seq2Seq}           & \textbf{74.13}       & 24.11       & 52.39          & \textbf{68.59}      & 20.35      & 46.53    & \textbf{62.25}       & 14.23      & 37.17     & \textbf{68.92}      & 19.68      & 45.94         \\
\textit{3$\times$Seq2Edit+3$\times$Seq2Seq}           & 72.82       & \textbf{26.38}       & \textbf{53.81}          & 67.95      & \textbf{21.58}      & \textbf{47.52}    & 60.65       & \textbf{16.39}      & \textbf{39.38}     & 67.76      & \textbf{21.42}      & \textbf{47.29}         \\
\midrule
Human              & 75.77       & 66.15       & 73.63        & 74.14       & 64.84      & 72.00          & 72.31      & 62.26      & 70.05          & 73.47      & 63.75      & 71.25         \\ 
\bottomrule
\end{tabular}
}
\caption{Performance of models and our annotators on MuCGEC, using the \textbf{char-based} metric. 
For calculating the human performance, 
each submitted result is considered as a sample if an annotator submits multiple references. 
}
\label{tab:benchmark:ours}
\end{table*}

%% file: tables/error_type_eval.tex
\begin{table}
\centering
\scalebox{0.71}{
\begin{tabular}{lcccc}
\toprule
\textbf{}        & \textbf{\textit{Seq2Edit}}  & \textbf{\textit{Seq2Seq}} & \textbf{\textit{Ensemble}} & \textbf{Human} \\ \hline
\uline{M}issing (\textbf{29.2\%})                  & 41.09 & 40.93 & 42.25             & 69.72          \\
\uline{R}edundant (\textbf{16.1\%})                & 43.11 & 37.65 & 54.18             & 72.78          \\
\uline{S}ubstitution (\textbf{48.9\%})                  & 35.99  & 39.98  & 47.37            & 71.69          \\
\uline{W}ord-order (\textbf{5.8\%})                & 28.28 & 40.33   & 42.44             & 72.58          \\ \bottomrule
\end{tabular}
}
\caption{$F_{0.5}$ scores for each  error type on All-test. The bold numbers in parentheses show the proportion of each error type. ``{\textit{Ensemble}}'' refers to  ``{\textit{3$\times$Seq2Edit+3$\times$Seq2Seq}}''.}
\label{tab:type:eval}
\end{table}

%% file: figure/answer_num.tex
\begin{figure}[tb!] %
\centering %
\scalebox{0.64}{
\begin{tikzpicture}
\begin{axis}[
    legend style={
                draw=none,
                fill opacity=2,
                legend cell align={left},
                text=black,
                at={(1.01,0.99)},
                anchor=north west,
                font=\small,
              },
    xlabel={Maximum Number of References},
    ylabel={$F_{0.5}$},
    ylabel style = {yshift=-10pt},
    ymin=20, ymax=80,
    symbolic x coords={1,2,3,All},
    xtick=data,
    ytick={20,40,60,80},
    ymajorgrids=true,
    xmajorgrids=true,
    grid style=dashed,
]

\addplot[
    color=midnightblue!80,dashdotted,thick,
    mark=square*,
    mark options={solid,mark size=2pt}
    ]
    coordinates {
    (1,48.17)(2,64.21)(3,70.01)(All,71.25)
    };
    \addlegendentry{Human}
    
\addplot[
    dashed,thick,color=forestgreen,
    mark=diamond*,
    mark options={solid,mark size=2pt},
    ]
    coordinates {
    (1,29.17)(2,36.46)(3,38.80)(All,39.22)
    };
    \addlegendentry{Seq2Edit}
    
\addplot[
    thick,color=brickred!80,
    mark=triangle*,
    mark options={solid,mark size=3pt},
    ]
    coordinates {
    (1,29.64)(2,36.95)(3,39.16)(All,39.58)
    };
    \addlegendentry{Seq2Seq}
    
\addplot[
    densely dashed, thick,color=purple!80,
    mark=*,
    mark options={solid,mark size=2pt},
    ]
    coordinates {
    (1,32.97)(2,42.50)(3,45.52)(All,45.94)
    };
    \addlegendentry{1$\times$S2E+1$\times$S2S}
    
\addplot[
    densely dashed, thick,color=burntorange!80,
    mark=pentagon*,
    mark options={solid,mark size=3pt},
    ]
    coordinates {
    (1,33.49)(2,43.69)(3,46.83)(All,47.29)
    };
    \addlegendentry{3$\times$S2E+3$\times$S2S}
    
\end{axis}
\end{tikzpicture}
}

\caption{%
Effect of the number of references on $F_{0.5}$.
}
\label{fig:answer:num}
\end{figure}
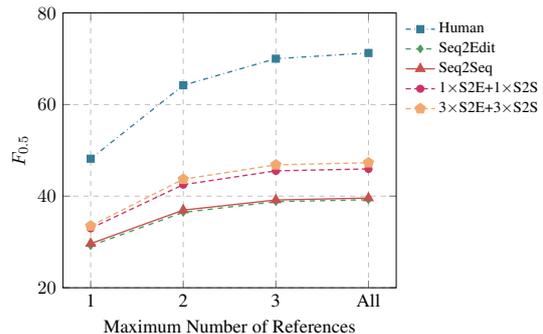

%% file: chapters/related_work.tex
\section{Related Works}

\textbf{EGEC resources.} 
There has been a lot of work on 
EGEC data construction. 
As the two earliest EGEC datasets, FCE \citep{yannakoudakis2011new} and NUCLE \citep{dahlmeier2013building} %
adopt the \emph{error-coded} annotation paradigm.  %
In contrast, JFLEG \citep{napoles2017jfleg} collects sentences from TOFEL exams and adopts the \emph{direct rewriting} paradigm. %
W\&I \citep{bryant2019bea} also chooses the \emph{direct rewriting} paradigm, and %
extra annotates a score indicating the language proficiency level of the writer for each input sentence. 
All four datasets  
are composed of essays from non-native English speakers and provide multiple references.

Recently, researchers have started to annotate small-scale EGEC data for texts written by  
native English speakers, including AESW  \citep{daudaravicius2016report}, LOCNESS  \citep{bryant2019bea}, GMEG  \citep{napoles2019enabling} and CWEB \citep{flachs2020grammatical}. %
In the future, we plan to extend this work to texts written by native Chinese speakers.

\textbf{CGEC resources.} 
Compared with EGEC, 
progress in CGEC data construction largely lags behind.  
As discussed in Section \ref{sec:intro}, NLPCC18 \citep{zhao2018overview} and CGED \citep{rao2018overview,rao2020overview} are the only two evaluation datasets for CGEC research. Besides them, there are also a few resources for training CGEC models, e.g., the Lang8 corpus \citep{zhao2018overview} and the HSK corpus \citep{zhang2009hsk}. 

Concurrently with this work, \citet{wang2022yaclc} present a multi-reference CGEC dataset, named as yet another Chinese learner corpus (YACLC), containing 32,124 sentences %
from Lang8.  
Each sentence is annotated by 10 annotators.

\textbf{Recent progress in CGEC.} 
In the NLPCC-2018 shared task \citep{zhao2018overview}, many systems adopt Seq2Seq models, based on RNN/CNN.
Recent work mainly utilizes Transformer \citep{wang2020base, zhao2020maskgec, tang2021WCDA}. %
\citet{hinson2020heterogeneous} first employ a Seq2Edit model for CGEC, and achieve comparable performance with the Seq2Seq counterparts. Some systems in the CGED-2020 shared task  \citep{rao2020overview} directly employ the open-source Seq2Edit model, i.e., GECToR \citep{liang2020bert}.
Most Seq2Edit models use PLMs like BERT \citep{devlin2019bert} to initialize their encoders. 
Besides the above two mainstream models,  \citet{DBLP:conf/acl/Li020} for the first time apply a non-autoregressive neural machine translation model to CGEC. 

Besides modeling optimization, %
techniques like
data augmentation \citep{zhao2020maskgec,tang2021WCDA} and model ensemble \citep{hinson2020heterogeneous} have proven to  be very useful for CGEC.

%% file: chapters/conclusion.tex
\section{Conclusions}

This paper presents MuCGEC, a newly annotated evaluation dataset for CGEC, consisting of 7,063 sentences written by CSL learners. Compared with existing CGEC datasets, ours can support more reliable evaluation due to three important features: 1) providing multiple references; 2) covering three  text sources; 3) adopting strict quality control (i.e., annotation guidelines and workflow). 
After describing the data construction process, we perform detailed analyses of our data. %
Then, we adopt two mainstream and competitive CGEC models, i.e., Seq2Seq and Seq2Edit, and carry out benchmark experiments. 
We also propose to adopt char-based evaluation metrics to 
replace previously used word-based ones. %

\section*{Acknowledgements}
We want to thank the anonymous reviewers and Sebastian Schuster for their great help. We are also grateful to Zeyang Liu for building the annotation system, and Lei Zhang, Fukang Yan, Jiayu Shen, Houquan Zhou, Yu Zhang for helping us improve this paper, and all annotators for their great effort in data annotation. This work was partially supported by National Natural Science Foundation of China (Grant No.62176173 and No.61876116) and by Alibaba Group through Alibaba Innovative  Research Program.

%% file: chapters/appendix.tex
\appendix

\section{Interface}

\label{sec:interface}

Figure \ref{fig:anno:interface} shows our design of annotation interface in our annotation tool, where annotators correct assigned sentences. 
Given an annotation task, this interface presents a potentially wrong sentence and a text input box. The original sentence is copied into the input box below, so that the annotator can directly modify it. To support multiple corrections, we also provide a button  to add additional input boxes. Two special buttons are provided to deal with special cases. The \emph{error free} button means that the sentence is correct; the \emph{not annotatable} button means that the annotator can not understand the sentence. 

\input{figure/annotation_interface}

Figure \ref{fig:review:interface} shows the review interface, where senior annotators judge whether the submitted corrections are correct. All corrections of a sentence from annotators are  shown on the screen, and reviewers click a check box to mark each of them as correct or incorrect. The input box below allows reviewers to supplement extra valid corrections.

\input{figure/review_interface}

\section{Hyperparameters}

\label{sec:hp}

Table \ref{tab:hyper} and Table \ref{tab:hyper:s2e} shows the detailed hyperparameters for training our two benchmark models. 
Due to the GPU memory limitation, we truncated sentences longer than 100 characters when training the Seq2Seq model. In other words, extra characters in the input sentences and the references are discarded.

\textbf{A useful trick.} We find that some sentences in MuCGEC are actually composed of multiple sentences. Therefore, we split one input sentence into multiple ones based on punctuation marks such as periods, exclamation marks, and so on. Then, we perform corrections on the smaller sentences and concatenate the results. 
We find this trick can consistently improve performance in our preliminary experiments. 
For now, we decide not to break the sentences when releasing in order to be consistent with the sources where the data comes from. 

 \input{tables/hypermeters}
 
 \input{tables/seq2edit}

\section{More Discussion on the Char-based Span-level Metric}
We find that ERRANT\_ZH \citep{hinson2020heterogeneous}, a useful evaluation tool for CGEC, only merges edits for redundant errors and missing errors, and does not merge edits for substitution errors. However, as discussed in Section \ref{sec:eval-metric}, we also merge the consecutive edits for substitution errors. We hope future research can adopt our simplified version unless there is a strong reason. Meanwhile, we should keep thinking about which evaluation metrics are more suitable for CGEC task.

%% file: figure/annotation_interface.tex
\begin{center}
\begin{figure}[htp!] %
\centering %
\includegraphics[height=3cm, width=8cm]{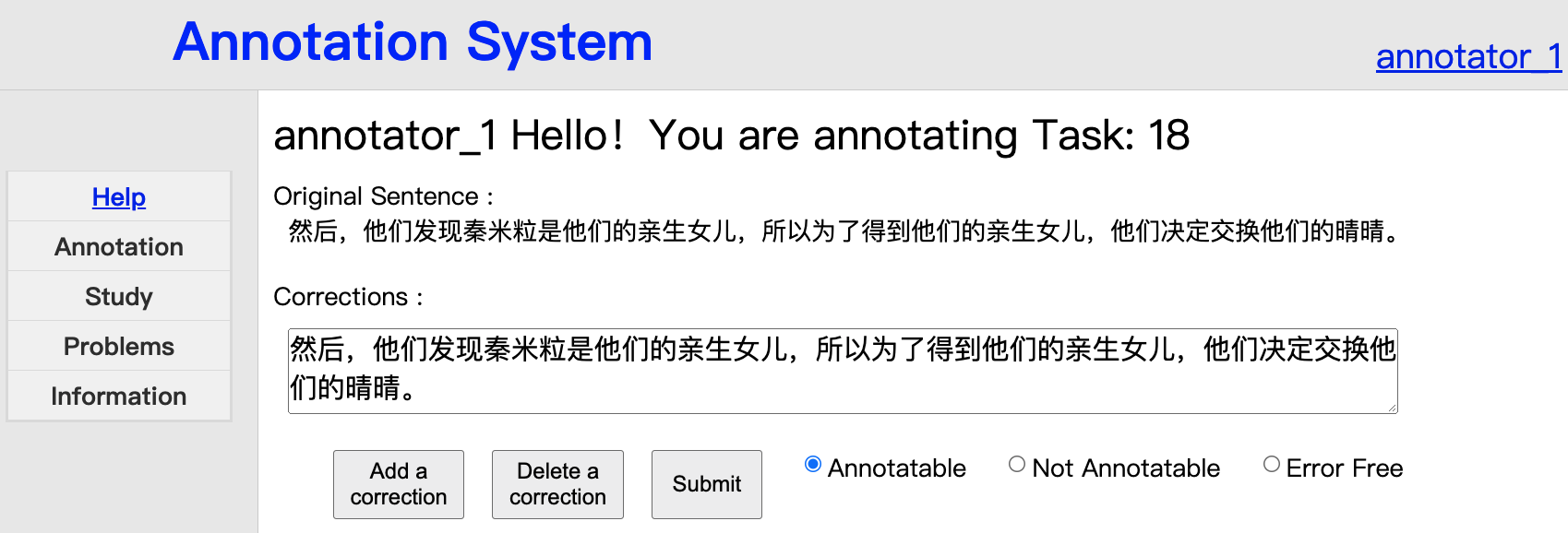}
\caption{The screenshot of the annotation interface.}
\label{fig:anno:interface}
\end{figure}

\end{center}

%% file: figure/review_interface.tex
\begin{center}
\begin{figure}[!htp] %
\centering %
\includegraphics[height=3cm, width=8cm]{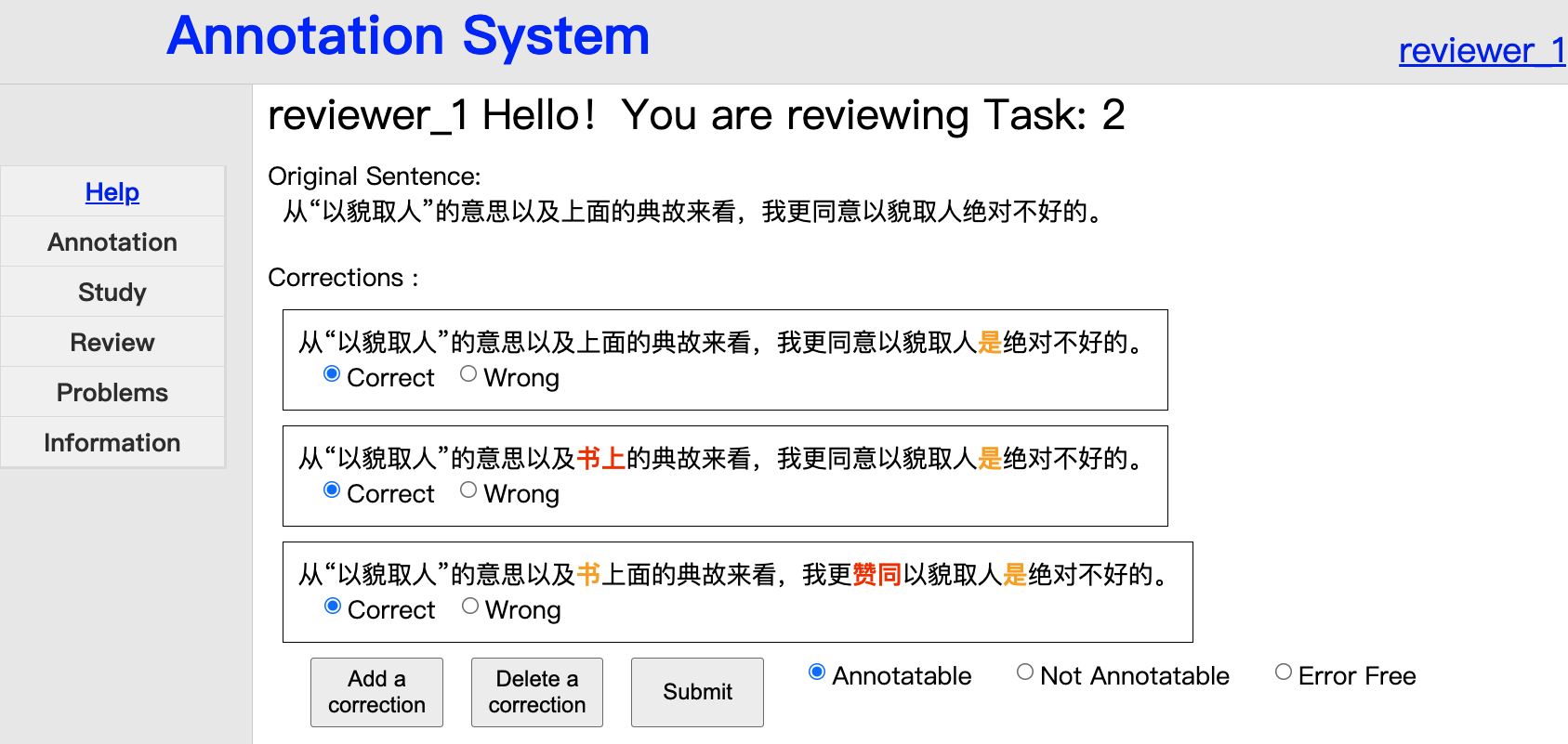}
\caption{The screenshot of the review interface.}
\label{fig:review:interface}
\end{figure}
\end{center}

%% file: tables/hypermeters.tex
\begin{center}
\begin{table}[!thp]
\centering
\scalebox{0.66}{
\begin{tabular}{cc}
\toprule
\textbf{Configurations}   & \textbf{Values}                                   \\ \hline
Model architecture        & BART \citep{lewis2020bart}                               \\
Pretrained model       & Chinese-BART-Large \citep{shao2021cpt}                        \\
Number of max epochs          & 10                                                \\
Devices                   & 8 Nvidia V100 GPU (32GB)                                    \\
Batch size per GPU        & 32                                              \\
Optimizer                 & \begin{tabular}[c]{@{}c@{}}Adam \citep{kingma2014adam}  \\ ($\beta_1=0.9,\beta_2=0.999,\epsilon=1 \times 10^{-8}$) \end{tabular}      \\
Learning rate             & $3 \times 10^{-6}$                                           \\
Learning rate scheduler   & Polynomial                                      \\
Gradient accumulation steps   & 4                                      \\
Dropout                   & 0.1                                               \\
Gradient clipping                   & 1.0                                               \\
Loss function             & \begin{tabular}[c]{@{}c@{} }Label smoothed cross entropy \\ (label-smoothing=0.1) \\ \citep{szegedy2016rethinking} \end{tabular}\\ 
Total training time                         & About 20 hours \\
Stopping criteria         & Loss value on the dev set          \\
Patience                  & 3      \\
\bottomrule
\end{tabular}
}
\caption{Hyperparameter values of our Seq2Seq model.}
\label{tab:hyper}
\end{table}
\end{center}

%% file: tables/seq2edit.tex
\vspace{-1cm}
\begin{center}
\begin{table}[!thp]
\centering
\scalebox{0.66}{
\begin{tabular}{cc}
\toprule
\textbf{Configurations}   & \textbf{Values}                                   \\ \hline
Model architecture        & GECToR   \citep{omelianchuk2020gector}                                         \\
Pretrained model       & Chinese-Struct-Bert-Large \citep{wang2019structbert}                        \\
Number of max epochs      & 20                                                \\
Number of cold epochs     & 2                                                 \\
Devices                   & 1 Nvidia V100 GPU (32GB)  
         \\
Optimizer                   & \begin{tabular}[c]{@{}c@{}}Adam \citep{kingma2014adam}  \\ ($\beta_1=0.9,\beta_2=0.999,\epsilon=1 \times 10^{-8}$) \end{tabular}
         \\
Cold learning rate        & $1 \times 10^{-3}$                                             \\
Learning rate             & $1 \times 10^{-5}$                                               \\
Batch size                & 128                                               \\
Loss function             & Cross entropy \\ 
Total training time                         & About 10 hours \\
Stopping criteria         & Label prediction accuracy on the dev set          \\
Patience                  & 3      \\   
\bottomrule
\end{tabular}
}
\caption{Hyperparameter values of our Seq2Edit model. ``Cold'' means that freeze the parameters of BERT.}
\label{tab:hyper:s2e}
\end{table}
\end{center}

%% file: acl.bbl
\begin{thebibliography}{52}
\expandafter\ifx\csname natexlab\endcsname\relax\def\natexlab#1{#1}\fi

\bibitem[{Awasthi et~al.(2019)Awasthi, Sarawagi, Goyal, Ghosh, and
  Piratla}]{awasthi2019parallel}
Abhijeet Awasthi, Sunita Sarawagi, Rasna Goyal, Sabyasachi Ghosh, and Vihari
  Piratla. 2019.
\newblock Parallel iterative edit models for local sequence transduction.
\newblock In \emph{Proceedings of EMNLP-IJCNLP}, pages 4260--4270.

\bibitem[{Bryant et~al.(2019)Bryant, Felice, Andersen, and
  Briscoe}]{bryant2019bea}
Christopher Bryant, Mariano Felice, {\O}istein~E Andersen, and Ted Briscoe.
  2019.
\newblock The {BEA}-2019 shared task on grammatical error correction.
\newblock In \emph{Proceedings of BEA@ACL}, pages 52--75.

\bibitem[{Bryant et~al.(2017)Bryant, Felice, and Briscoe}]{bryant2017automatic}
Christopher Bryant, Mariano Felice, and Ted Briscoe. 2017.
\newblock Automatic annotation and evaluation of error types for grammatical
  error correction.
\newblock In \emph{Proceedings of ACL}, pages 793--805.

\bibitem[{Bryant and Ng(2015)}]{bryant2015far}
Christopher Bryant and Hwee~Tou Ng. 2015.
\newblock How far are we from fully automatic high quality grammatical error
  correction?
\newblock In \emph{Proceedings of ACL}, pages 697--707.

\bibitem[{Choshen and Abend(2018)}]{choshen2018inherent}
Leshem Choshen and Omri Abend. 2018.
\newblock Inherent biases in reference-based evaluation for grammatical error
  correction.
\newblock In \emph{Proceedings of ACL}, pages 632--642.

\bibitem[{Cui et~al.(2020)Cui, Che, Liu, Qin, Wang, and Hu}]{cui2020revisiting}
Yiming Cui, Wanxiang Che, Ting Liu, Bing Qin, Shijin Wang, and Guoping Hu.
  2020.
\newblock Revisiting pre-trained models for {Chinese} natural language
  processing.
\newblock In \emph{Proceedings of EMNLP: findings}, pages 657--668.

\bibitem[{Dahlmeier and Ng(2012)}]{dahlmeier2012better}
Daniel Dahlmeier and Hwee~Tou Ng. 2012.
\newblock Better evaluation for grammatical error correction.
\newblock In \emph{Proceedings of NAACL-HLT}, pages 568--572.

\bibitem[{Dahlmeier et~al.(2013)Dahlmeier, Ng, and Wu}]{dahlmeier2013building}
Daniel Dahlmeier, Hwee~Tou Ng, and Siew~Mei Wu. 2013.
\newblock Building a large annotated corpus of learner {English}: The nus
  corpus of learner {English}.
\newblock In \emph{Proceedings of BEA@NAACL-HLT}, pages 22--31.

\bibitem[{Daudaravicius et~al.(2016)Daudaravicius, Banchs, Volodina, and
  Napoles}]{daudaravicius2016report}
Vidas Daudaravicius, Rafael~E Banchs, Elena Volodina, and Courtney Napoles.
  2016.
\newblock A report on the automatic evaluation of scientific writing shared
  task.
\newblock In \emph{Proceedings of BEA@NAACL-HLT}, pages 53--62.

\bibitem[{Devlin et~al.(2019)Devlin, Chang, Lee, and
  Toutanova}]{devlin2019bert}
Jacob Devlin, Ming-Wei Chang, Kenton Lee, and Kristina Toutanova. 2019.
\newblock {BERT}: Pre-training of deep bidirectional transformers for language
  understanding.
\newblock In \emph{Proceedings of NAACL-HLT}, pages 4171--4186.

\bibitem[{Felice et~al.(2016)Felice, Bryant, and Briscoe}]{felice2016automatic}
Mariano Felice, Christopher Bryant, and Ted Briscoe. 2016.
\newblock Automatic extraction of learner errors in {ESL} sentences using
  linguistically enhanced alignments.
\newblock In \emph{Proceedings of COLING}, pages 825--835.

\bibitem[{Flachs et~al.(2020)Flachs, Lacroix, Yannakoudakis, Rei, and
  S{\o}gaard}]{flachs2020grammatical}
Simon Flachs, Oph{\'e}lie Lacroix, Helen Yannakoudakis, Marek Rei, and Anders
  S{\o}gaard. 2020.
\newblock Grammatical error correction in low error density domains: a new
  benchmark and analyses.
\newblock In \emph{Proceedings of EMNLP}, pages 8467--8478.

\bibitem[{Fu et~al.(2020)Fu, Liu, Zhang, and Huang}]{fu2020chinese}
Jinlan Fu, Pengfei Liu, Qi~Zhang, and Xuan-Jing Huang. 2020.
\newblock Is {Chinese} word segmentation a solved task? {Rethinking} neural
  {Chinese} word segmentation.
\newblock In \emph{Proceedings of EMNLP}, pages 5676--5686.

\bibitem[{Fu et~al.(2018)Fu, Huang, and Duan}]{fu2018youdao}
Kai Fu, Jin Huang, and Yitao Duan. 2018.
\newblock Youdao’s winning solution to the {NLPCC}-2018 task 2 challenge: a
  neural machine translation approach to {Chinese} grammatical error
  correction.
\newblock In \emph{CCF International Conference on Natural Language Processing
  and Chinese Computing (NLPCC)}, pages 341--350.

\bibitem[{Grundkiewicz et~al.(2020)Grundkiewicz, Bryant, and
  Felice}]{grundkiewicz2020crash}
Roman Grundkiewicz, Christopher Bryant, and Mariano Felice. 2020.
\newblock A crash course in automatic grammatical error correction.
\newblock In \emph{Proceedings of COLING: Tutorial Abstracts}, pages 33--38.

\bibitem[{Hinson et~al.(2020)Hinson, Huang, and Chen}]{hinson2020heterogeneous}
Charles Hinson, Hen-Hsen Huang, and Hsin-Hsi Chen. 2020.
\newblock Heterogeneous recycle generation for {Chinese} grammatical error
  correction.
\newblock In \emph{Proceedings of COLING}, pages 2191--2201.

\bibitem[{Junczys-Dowmunt et~al.(2018)Junczys-Dowmunt, Grundkiewicz, Guha, and
  Heafield}]{junczys2018approaching}
Marcin Junczys-Dowmunt, Roman Grundkiewicz, Shubha Guha, and Kenneth Heafield.
  2018.
\newblock Approaching neural grammatical error correction as a low-resource
  machine translation task.
\newblock In \emph{Proceedings of NAACL-HLT}, pages 595--606.

\bibitem[{Katsumata and Komachi(2020)}]{katsumata2020stronger}
Satoru Katsumata and Mamoru Komachi. 2020.
\newblock Stronger baselines for grammatical error correction using a
  pretrained encoder-decoder model.
\newblock In \emph{Proceedings of AACL}, pages 827--832.

\bibitem[{Kingma and Ba(2014)}]{kingma2014adam}
Diederik~P Kingma and Jimmy Ba. 2014.
\newblock Adam: a method for stochastic optimization.
\newblock \emph{arXiv preprint arXiv:1412.6980}.

\bibitem[{Lewis et~al.(2020)Lewis, Liu, Goyal, Ghazvininejad, Mohamed, Levy,
  Stoyanov, and Zettlemoyer}]{lewis2020bart}
Mike Lewis, Yinhan Liu, Naman Goyal, Marjan Ghazvininejad, Abdelrahman Mohamed,
  Omer Levy, Veselin Stoyanov, and Luke Zettlemoyer. 2020.
\newblock {BART}: denoising sequence-to-sequence pre-training for natural
  language generation, translation, and comprehension.
\newblock In \emph{Proceedings of ACL}, pages 7871--7880.

\bibitem[{Li and Shi(2021)}]{DBLP:conf/acl/Li020}
Piji Li and Shuming Shi. 2021.
\newblock Tail-to-tail non-autoregressive sequence prediction for {Chinese}
  grammatical error correction.
\newblock In \emph{Proceedings of ACL}, pages 4973--4984.

\bibitem[{Liang et~al.(2020)Liang, Zheng, Guo, Cui, Xiong, Rong, and
  Dong}]{liang2020bert}
Deng Liang, Chen Zheng, Lei Guo, Xin Cui, Xiuzhang Xiong, Hengqiao Rong, and
  Jinpeng Dong. 2020.
\newblock {BERT} enhanced neural machine translation and sequence tagging model
  for {Chinese} grammatical error diagnosis.
\newblock In \emph{Proceedings of NLPTEA@AACL}, pages 57--66.

\bibitem[{Liu et~al.(2019)Liu, Ott, Goyal, Du, Joshi, Chen, Levy, Lewis,
  Zettlemoyer, and Stoyanov}]{liu2019roberta}
Yinhan Liu, Myle Ott, Naman Goyal, Jingfei Du, Mandar Joshi, Danqi Chen, Omer
  Levy, Mike Lewis, Luke Zettlemoyer, and Veselin Stoyanov. 2019.
\newblock {RoBERTa}: a robustly optimized {BERT} pretraining approach.
\newblock \emph{arXiv preprint arXiv:1907.11692}.

\bibitem[{Malmi et~al.(2019)Malmi, Krause, Rothe, Mirylenka, and
  Severyn}]{malmi2019encode}
Eric Malmi, Sebastian Krause, Sascha Rothe, Daniil Mirylenka, and Aliaksei
  Severyn. 2019.
\newblock Encode, tag, realize: high-precision text editing.
\newblock In \emph{Proceedings of EMNLP-IJCNLP}, pages 5054--5065.

\bibitem[{Mita et~al.(2019)Mita, Mizumoto, Kaneko, Nagata, and
  Inui}]{mita2019cross}
Masato Mita, Tomoya Mizumoto, Masahiro Kaneko, Ryo Nagata, and Kentaro Inui.
  2019.
\newblock Cross-corpora evaluation and analysis of grammatical error correction
  models—is single-corpus evaluation enough?
\newblock In \emph{Proceedings of NAACL-HLT (Short)}, pages 1309--1314.

\bibitem[{Nagata and Sakaguchi(2016)}]{nagata2016phrase}
Ryo Nagata and Keisuke Sakaguchi. 2016.
\newblock Phrase structure annotation and parsing for learner {English}.
\newblock In \emph{Proceedings of ACL}, pages 1837--1847.

\bibitem[{Napoles et~al.(2019)Napoles, N{\u{a}}dejde, and
  Tetreault}]{napoles2019enabling}
Courtney Napoles, Maria N{\u{a}}dejde, and Joel Tetreault. 2019.
\newblock Enabling robust grammatical error correction in new domains: data
  sets, metrics, and analyses.
\newblock \emph{TACL}, 7:551--566.

\bibitem[{Napoles et~al.(2017)Napoles, Sakaguchi, and
  Tetreault}]{napoles2017jfleg}
Courtney Napoles, Keisuke Sakaguchi, and Joel Tetreault. 2017.
\newblock {JFLEG}: a fluency corpus and benchmark for grammatical error
  correction.
\newblock In \emph{Proceedings of EACL}, pages 229--234.

\bibitem[{Ng et~al.(2014)Ng, Wu, Briscoe, Hadiwinoto, Susanto, and
  Bryant}]{ng2014conll}
Hwee~Tou Ng, Siew~Mei Wu, Ted Briscoe, Christian Hadiwinoto, Raymond~Hendy
  Susanto, and Christopher Bryant. 2014.
\newblock The {CoNLL}-2014 shared task on grammatical error correction.
\newblock In \emph{Proceedings of CoNLL: Shared Task}, pages 1--14.

\bibitem[{Ng et~al.(2013)Ng, Wu, Wu, Hadiwinoto, and Tetreault}]{ng2013conll}
Hwee~Tou Ng, Siew~Mei Wu, Yuanbin Wu, Christian Hadiwinoto, and Joel Tetreault.
  2013.
\newblock The {CoNLL}-2013 shared task on grammatical error correction.
\newblock In \emph{Proceedings of CoNLL: Shared Task}, pages 1--12.

\bibitem[{Náplava et~al.(2022)Náplava, Straka, Straková, and
  Rosen}]{10.1162/tacl_a_00470}
Jakub Náplava, Milan Straka, Jana Straková, and Alexandr Rosen. 2022.
\newblock Czech grammar error correction with a large and diverse corpus.
\newblock \emph{TACL}, 10:452--467.

\bibitem[{Omelianchuk et~al.(2020)Omelianchuk, Atrasevych, Chernodub, and
  Skurzhanskyi}]{omelianchuk2020gector}
Kostiantyn Omelianchuk, Vitaliy Atrasevych, Artem Chernodub, and Oleksandr
  Skurzhanskyi. 2020.
\newblock Gector--grammatical error correction: tag, not rewrite.
\newblock In \emph{Proceedings of BEA@ACL}, pages 163--170.

\bibitem[{Rao et~al.(2018)Rao, Gong, Zhang, and Xun}]{rao2018overview}
Gaoqi Rao, Qi~Gong, Baolin Zhang, and Endong Xun. 2018.
\newblock Overview of {NLPTEA}-2018 share task {Chinese} grammatical error
  diagnosis.
\newblock In \emph{Proceedings of NLPTEA@ACL}, pages 42--51.

\bibitem[{Rao et~al.(2020)Rao, Yang, and Zhang}]{rao2020overview}
Gaoqi Rao, Erhong Yang, and Baolin Zhang. 2020.
\newblock Overview of {NLPTEA}-2020 shared task for {Chinese} grammatical error
  diagnosis.
\newblock In \emph{Proceedings of NLPTEA@AACL}, pages 25--35.

\bibitem[{Ren et~al.(2018)Ren, Yang, and Xun}]{ren2018sequence}
Hongkai Ren, Liner Yang, and Endong Xun. 2018.
\newblock A sequence to sequence learning for {Chinese} grammatical error
  correction.
\newblock In \emph{CCF International Conference on Natural Language Processing
  and Chinese Computing (NLPCC)}, pages 401--410.

\bibitem[{Rothe et~al.(2021)Rothe, Mallinson, Malmi, Krause, and
  Severyn}]{rothe2021recipe}
Sascha Rothe, Jonathan Mallinson, Eric Malmi, Sebastian Krause, and Aliaksei
  Severyn. 2021.
\newblock A simple recipe for multilingual grammatical error correction.
\newblock In \emph{Proceedings of ACL-IJCNLP}, pages 702--707.

\bibitem[{Sakaguchi et~al.(2016)Sakaguchi, Napoles, Post, and
  Tetreault}]{sakaguchi2016reassessing}
Keisuke Sakaguchi, Courtney Napoles, Matt Post, and Joel Tetreault. 2016.
\newblock Reassessing the goals of grammatical error correction: fluency
  instead of grammaticality.
\newblock \emph{TACL}, 4:169--182.

\bibitem[{Shao et~al.(2021)Shao, Geng, Liu, Dai, Yang, Zhe, Bao, and
  Qiu}]{shao2021cpt}
Yunfan Shao, Zhichao Geng, Yitao Liu, Junqi Dai, Fei Yang, Li~Zhe, Hujun Bao,
  and Xipeng Qiu. 2021.
\newblock {CPT}: a pre-trained unbalanced transformer for both {Chinese}
  language understanding and generation.
\newblock \emph{arXiv preprint arXiv:2109.05729}.

\bibitem[{Syvokon and Nahorna(2021)}]{syvokon2021ua}
Oleksiy Syvokon and Olena Nahorna. 2021.
\newblock {UA-GEC}: grammatical error correction and fluency corpus for the
  {Ukrainian} language.
\newblock \emph{arXiv preprint arXiv:2103.16997}.

\bibitem[{Szegedy et~al.(2016)Szegedy, Vanhoucke, Ioffe, Shlens, and
  Wojna}]{szegedy2016rethinking}
Christian Szegedy, Vincent Vanhoucke, Sergey Ioffe, Jon Shlens, and Zbigniew
  Wojna. 2016.
\newblock Rethinking the inception architecture for computer vision.
\newblock In \emph{Proceedings of ICCV}, pages 2818--2826.

\bibitem[{Tang et~al.(2021)Tang, Ji, Zhao, and Li}]{tang2021WCDA}
Zecheng Tang, Yixin Ji, Yibo Zhao, and Junhui Li. 2021.
\newblock {Chinese} grammatical error correction enhanced by data augmentation
  from word and character levels.
\newblock In \emph{Proceedings of the 20th Chinese National Conference on
  Computational Linguistics (CCL) (in {Chinese})}, pages 813--824.

\bibitem[{Vaswani et~al.(2017)Vaswani, Shazeer, Parmar, Uszkoreit, Jones,
  Gomez, Kaiser, and Polosukhin}]{vaswani2017attention}
Ashish Vaswani, Noam Shazeer, Niki Parmar, Jakob Uszkoreit, Llion Jones,
  Aidan~N Gomez, {\L}ukasz Kaiser, and Illia Polosukhin. 2017.
\newblock Attention is all you need.
\newblock In \emph{Proceedings of NIPS}, pages 5998--6008.

\bibitem[{Wang et~al.(2020)Wang, Yang, Wang, Du, and Yang}]{wang2020base}
Chencheng Wang, Liner Yang, Yingying Wang, Yongping Du, and Erhong Yang. 2020.
\newblock {Chinese} grammatical error correction method based on transformer
  enhanced architecture.
\newblock \emph{Journal of Chinese Information Processing (in {Chinese})},
  34(6):106--114.

\bibitem[{Wang et~al.(2019)Wang, Bi, Yan, Wu, Xia, Bao, Peng, and
  Si}]{wang2019structbert}
Wei Wang, Bin Bi, Ming Yan, Chen Wu, Jiangnan Xia, Zuyi Bao, Liwei Peng, and
  Luo Si. 2019.
\newblock {StructBERT}: incorporating language structures into pre-training for
  deep language understanding.
\newblock In \emph{Proceedings of ICLR}.

\bibitem[{Wang et~al.(2022)Wang, Kong, Yang, Wang, Lu, Hu, He, Liu, Chen, Yang,
  and Sun}]{wang2022yaclc}
Yingying Wang, Cunliang Kong, Liner Yang, Yijun Wang, Xiaorong Lu, Renfen Hu,
  Shan He, Zhenghao Liu, Yun Chen, Erhong Yang, and Maosong Sun. 2022.
\newblock {YACLC}: a {Chinese} learner corpus with multidimensional annotation.
\newblock \emph{arXiv preprint arXiv:2104.05507}.

\bibitem[{Wang et~al.(2021)Wang, Wang, Dang, Liu, and
  Liu}]{wang2021comprehensive}
Yu~Wang, Yuelin Wang, Kai Dang, Jie Liu, and Zhuo Liu. 2021.
\newblock A comprehensive survey of grammatical error correction.
\newblock \emph{ACM Transactions on Intelligent Systems and Technology (TIST)},
  12(5):1--51.

\bibitem[{Yannakoudakis et~al.(2011)Yannakoudakis, Briscoe, and
  Medlock}]{yannakoudakis2011new}
Helen Yannakoudakis, Ted Briscoe, and Ben Medlock. 2011.
\newblock A new dataset and method for automatically grading {ESOL} texts.
\newblock In \emph{Proceedings of ACL}, pages 180--189.

\bibitem[{Yuan and Briscoe(2016)}]{yuan2016grammatical}
Zheng Yuan and Ted Briscoe. 2016.
\newblock Grammatical error correction using neural machine translation.
\newblock In \emph{Proceedings of NAACL-HLT}, pages 380--386.

\bibitem[{Zhang(2009)}]{zhang2009hsk}
Baolin Zhang. 2009.
\newblock Features and functions of the {HSK} dynamic composition corpus (in
  {Chinese}).
\newblock \emph{International Chinese Language Education}, 4:71--79.

\bibitem[{Zhao et~al.(2018)Zhao, Jiang, Sun, and Wan}]{zhao2018overview}
Yuanyuan Zhao, Nan Jiang, Weiwei Sun, and Xiaojun Wan. 2018.
\newblock Overview of the {NLPCC} 2018 shared task: grammatical error
  correction.
\newblock In \emph{CCF International Conference on Natural Language Processing
  and Chinese Computing (NLPCC)}, pages 439--445.

\bibitem[{Zhao and Wang(2020)}]{zhao2020maskgec}
Zewei Zhao and Houfeng Wang. 2020.
\newblock {MaskGEC}: improving neural grammatical error correction via dynamic
  masking.
\newblock In \emph{Proceedings of AAAI}, pages 1226--1233.

\bibitem[{Zhou et~al.(2018)Zhou, Li, Liu, Bao, Xu, and Li}]{zhou2018chinese}
Junpei Zhou, Chen Li, Hengyou Liu, Zuyi Bao, Guangwei Xu, and Linlin Li. 2018.
\newblock {Chinese} grammatical error correction using statistical and neural
  models.
\newblock In \emph{CCF International Conference on Natural Language Processing
  and Chinese Computing (NLPCC)}, pages 117--128.

\end{thebibliography}
